\documentclass{article}
\usepackage{ijcai22}

\usepackage{times}
\usepackage{soul}
\usepackage{url}
\usepackage[hidelinks]{hyperref}
\usepackage[utf8]{inputenc}
\usepackage[small]{caption}
\usepackage{graphicx}
\usepackage{amsmath}
\usepackage{amsthm}
\usepackage{booktabs}
\usepackage{algorithm}
\usepackage{algorithmic}
\usepackage{xspace}

\newcommand{\jsremove}[1]{} 

\newcommand{\IGNORE}[1]{{}}

\newcommand{\tablelabel}[2][\reflabel]{\label{table:#1-#2}}
\newcommand{\tableref}[2][\reflabel]{Table~\ref{table:#1-#2}}

\newcommand{\be}{\begin{equation}}
\newcommand{\ee}{\end{equation}}
\newcommand{\bea}{\begin{eqnarray}}
\newcommand{\eea}{\end{eqnarray}}
\newcommand{\beas}{\begin{eqnarray*}}
\newcommand{\eeas}{\end{eqnarray*}}

\newcommand{\ie}{i.e.~}
\newcommand{\eg}{e.g.~}

\newcommand{\etal}{et al.\xspace}

\newcommand{\cut}[1]{}

\usepackage{graphicx}
\usepackage{amsmath}
\usepackage{amssymb}
\usepackage{booktabs}
\usepackage{bbding}
\usepackage{color,xcolor}
\usepackage{multirow}
\usepackage{float}
\usepackage{adjustbox}
\usepackage{url}
\usepackage{hyperref}
\usepackage{subcaption}

\newcommand{\gen}{$G_{\phi_{g}}$}
\newcommand{\gend}{$D_{\theta_{d}}$ }
\newcommand{\gene}{$E_{\theta_{e}}$}
\newcommand{\cgene}{$E_{\theta_{c}}$}

\title{Contact2Grasp: 3D Grasp Synthesis via Hand-Object Contact Constraint}

\author{
Haoming Li$^1$
\and
Xinzhuo Lin$^1$\and
Yang Zhou$^{2}$\and
Xiang Li$^2$\and
Yuchi Huo$^3$\and
Jiming Chen$^1$\And
Qi Ye$^{1}$\footnote{Corresponding author.}
\affiliations
$^1$ Key Lab of CS$\&$AUS, Zhejiang University, Hangzhou, China\\
$^2$OPPO US Research Center, Palo Alto, USA\\
$^3$State Key Lab of CAD\&CG and Zhejiang Lab, Zhejiang University, Hangzhou, China
\emails
\{haomingli, linxinzhuo, cjm, qi.ye\}@zju.edu.cn, 
\{yang.zhou, xiang.li\}@oppo.com,\\
huo.yuchi.sc@gmail.com
}

\begin{document}
\maketitle

\begin{abstract}
3D grasp synthesis generates grasping poses given an input object. Existing works tackle the problem by learning a direct mapping from objects to the distributions of grasping poses. However, because the physical contact is sensitive to small changes in pose, the high-nonlinear mapping between 3D object representation to valid poses is considerably non-smooth, leading to poor generation efficiency and restricted generality. To tackle the challenge, we introduce an intermediate variable for grasp contact areas to constrain the grasp generation; in other words, we factorize the mapping into two sequential stages by assuming that grasping poses are fully constrained given contact maps: 1) we first learn contact map distributions to generate the potential contact maps for grasps; 2) then learn a mapping from the contact maps to the grasping poses. Further, we propose a penetration-aware optimization with the generated contacts as a consistency constraint for grasp refinement. Extensive validations on two public datasets show that our method outperforms state-of-the-art methods regarding grasp generation on various metrics.

\end{abstract}

\section{Introduction}

\begin{figure}[ht]
  \centering
    \includegraphics[width=\linewidth]{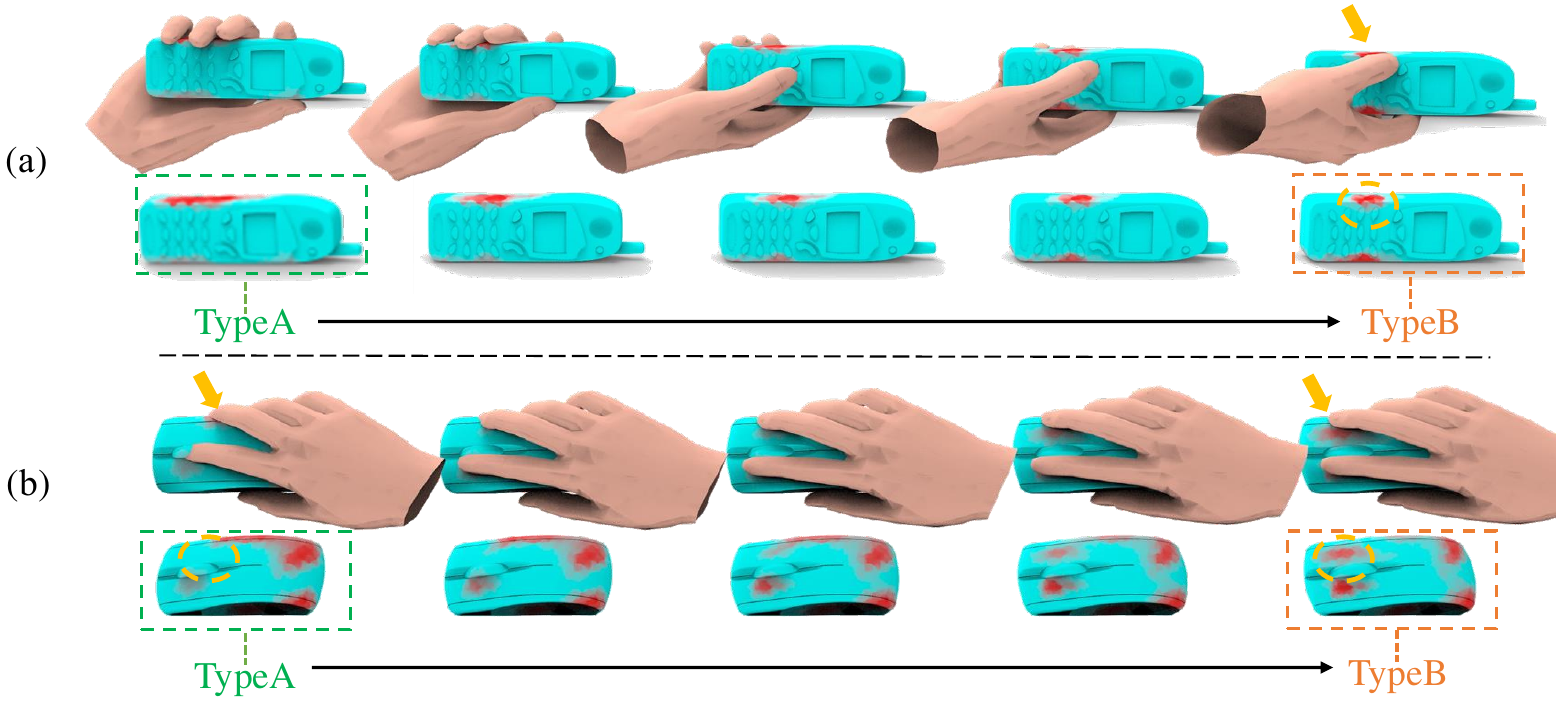}
  \caption{Interpolated contact maps and grasps between different generated contacts (TypeA and TypeB) by our method. Note that the grasping poses (\eg finger positions denoted in the yellow circle and arrow) change with transitions between two types of contacts and a small change in a valid contact map produces another valid grasp. The intermediate contact maps reduce the non-smooth high-nonlinear pose generation problem to a map generation problem in a low-dimension and smooth manifold, benefiting generation efficiency and generality.}
  \label{fig.contact_analysis}
  \vspace{-0.3cm}
\end{figure}

3D grasp synthesis studies the problem of generating grasping poses given an input object. It has wide applications ranging from animation, human-computer interaction to robotic grasping. Though it has been researched for many years, only a limited number of works about 3D grasp generation using deep learning have been proposed due to the lack of large grasping data ~\cite{corona2020ganhand,taheri2020grab,jiang2021hand,karunratanakul2020grasping,zhang2021manipnet,taheri2021goal}. Recently, a dataset for human grasping objects with annotations of full body meshes and objects meshes have been collected by a multi-view capture rig, and a coarse-to-fine hand pose generation network based on a conditional autoencoder (CVAE) is proposed ~\cite{taheri2020grab}. In ~\cite{karunratanakul2020grasping}, a new implicit representation is proposed for hand and object interactions, and a similar CVAE method is used for static grasps generation. Taheri \etal ~\cite{taheri2021goal} take a step further to learn dynamic grasping sequences including the whole body motion given an object, instead of static grasping poses.

Existing methods treat the generation as a black box mapping from an object to its grasp pose distribution. 
However, this formulation has its defects. On one hand, the mapping from the 3D object space to the pose space represented by rotations is highly non-linear. On the other hand, physical contact is sensitive to small changes in pose, e.g., less than a millimeter change in the pose of a fingertip normal to the surface of an object can make the difference between the object being held or dropped on the floor \cite{grady2021contactopt}. Therefore, the mapping between 3D object representation to valid poses is non-smooth, as a small change in the pose could make a valid pose invalid. These defects raise a challenge for the network to learn the sparse mapping and generalize to unseen valid poses in the highly non-linear space.

 \begin{figure*}[!htbp]
  \centering
  \vspace{-0.5cm}
    \includegraphics[width=0.9\textwidth]{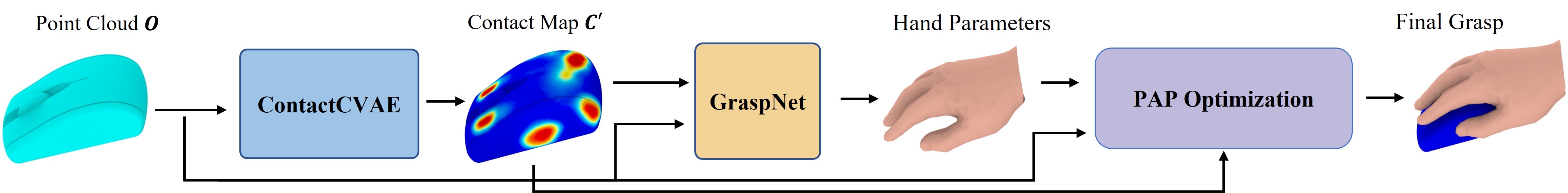}
  \caption{The framework of our method. It consists of three stages: ContactCVAE, GraspNet and Penetration-aware Partial Optimization. ContactCVAE takes an object point cloud $O$ as input and generates a contact map $C'$. GraspNet estimates a grasp parameterized by $\theta$ from the contact map $C'$.  Finally,  penetration-aware partial (PAP) optimization refines $\theta$ to get the final grasp.}
  \label{fig.figure1}
\end{figure*}

In robotics, contact areas between agents and objects are found to be important~\cite{deng20213d,roy2016multi,zhu2015understanding} because localizing the position of possible grasps can greatly help the planning of actions for robotic hands~\cite{mo2021where2act,wu2021vat,mandikal2021learning,mandikal2022dexvip}. For example, ~\cite{mo2021where2act} and~\cite{wu2021vat} first estimate the contact points for parallel-jaw grippers and plan paths to grasp the target objects. The common assumption in the literature is that the contact area is a point and the contact point generation is treated as a per-point (or pixel voxel) detection problem, \ie classifying each 3D object point to be a contact or not, which cannot be applied to dexterous hand grasps demonstrating much more complex contact. For dexterous robotic hand grasping, recent work~\cite{mandikal2021learning} finds that leveraging contact areas from human grasp can improve the grasping success rate in a reinforcement learning framework. However, it assumes an object only affords one grasp, which contradicts the real case and limits its application.

To tackle the limitations, we propose to leverage contact maps to constrain the grasp synthesis. Specifically, we factorize the learning task into two sequential stages, rather than taking a black-box hand pose generative network that directly maps an object to the possible grasping poses in previous work. In the first stage, we generate multiple hypotheses of the grasping contact areas, represented by binary 3D segmentation maps. In the second stage, we learn a mapping from the contact to the grasping pose by assuming the grasping pose is fully constrained given a contact map.

The intermediate segmentation contact maps align with the smooth manifold of the object surface: for example, a small change in a valid contact map would likely produce another valid solution (as illustrated in Figure~\ref{fig.contact_analysis}), then the corresponding pose can be deterministically established by the following GraspNet and PAP optimization. This manner reduces the challenging pose generation to an easier map generation problem in a low-dimension and smooth manifold, benefiting generation efficiency and generality.

The other benefit of the intermediate contact representation is enabling the optimization from the contacts. Different from the optimization for the full grasps from scratch~\cite{brahmbhatt2019contactgrasp,xing2022energy},  we propose a penetration-aware partial (PAP) optimization with the intermediate contacts. It detects partial poses causing penetration and leverages the generated contact maps as a consistency constraint for the refinement of the partial poses. The PAP optimization constrains gradients from wrong partial poses to affect these poses requiring adjustment only, which results in better grasp quality than a global optimization method.

In summary, our key contributions are: 1) we tackle the high non-linearity problem of the 3D generation problem by introducing the contact map constraint and factorizing the generation in two stages: contact map generation and mapping from contact maps to grasps; 2) we propose a PAP optimization with the intermediate contacts for the grasp refinement; 3) benefiting from the two decomposed learning stages and partial optimization, our method outperforms existing methods both quantitatively and qualitatively.

\section{Related Works}
Human grasp generation is a challenging task due to the higher degrees of freedom of human hands and the requirement of the generated hands to interact with objects in a physically reasonable manner. Most methods use models such as MANO~\cite{romero2017embodied} to parameterize hand poses, aiming to directly learn a latent conditional distribution of the hand parameters given objects via large datasets. The distribution is usually learned by generative network models such as Conditional Variational Auto-Encoder~\cite{sohn2015learning}, or Adversarial Generative Networks~\cite{arjovsky2017wasserstein}. To get finer poses, many existing works adopt a coarse-to-fine strategy by learning the residuals of the grasping poses in the refinement stage. \cite{corona2020ganhand} uses a generative adversarial network to obtain an initial grasp, and then an extra network to refine it. \cite{taheri2020grab} follows a similar strategy but uses a CVAE model to output an initial grasp.  

In recent works, contact maps are exploited to improve robotic grasping, hand object reconstruction, and 3D grasp synthesis. \cite{brahmbhatt2019contactgrasp} introduces a loss for robotic grasping optimization using contact maps captured from thermal cameras ~\cite{brahmbhatt2019contactdb,brahmbhatt2020contactpose} to filter and rank random grasps sampled by GraspIt!~\cite{miller2004graspit}. It concludes that synthesized grasping poses optimized directly from the contact demonstrate superior quality to other approaches which kinematically re-target observed human grasps to the target hand model. In the reconstruction of the hand-object interaction, \cite{grady2021contactopt} propose a differentiable contact optimization to refine the hand pose reconstructed from an image. In the 3D grasp synthesis, \cite{jiang2021hand} also exploits contact maps but they only use them to refine generated grasps during inference. 
\begin{figure*}[t]
  \centering
  \vspace{-0.4cm}
    \includegraphics[width=0.9\textwidth]{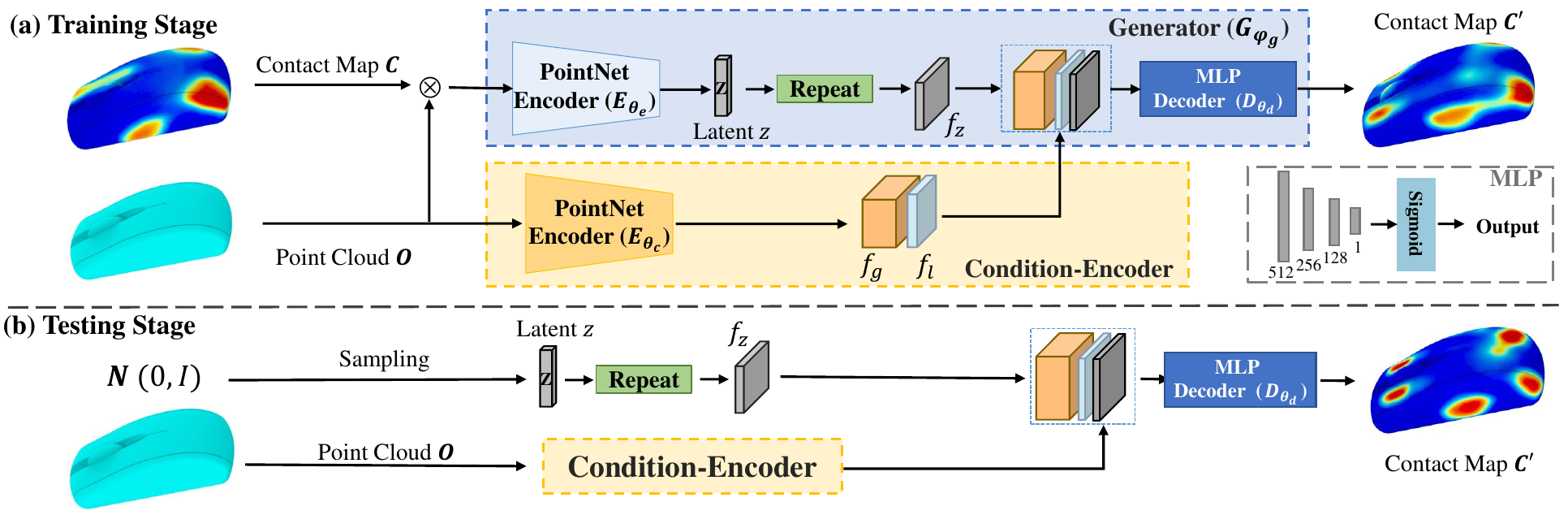}
  \caption{The architecture of ContactCVAE. (a) In the training stage, it takes both an object point cloud  and a contact map as input to reconstruct the contact map; (b) In the testing stage, by sampling from the latent distribution, it generates grasp contacts with an object point cloud as the conditional input only. $\otimes$ means concatenation.}
  \label{fig.ContactCVAE}
  \vspace{-0.4cm}
\end{figure*}
Our work differs from these works using contact maps in three aspects: 1) these works use contact maps as a loss for the grasp optimization or post-processing for further grasp refinement while our work exploits the contact maps as an intermediate constraint for the learning of the grasp distribution;  2) in contrast to the learning-based works with contact maps which treat objects-to-grasps as a black box, our work factorizes the grasp synthesis into objects-to-contact maps and contact maps-to-grasps; 3) moreover, these works refine the whole grasps with global optimization methods using contact maps while our penetration-aware partial optimization detects the partial poses causing the penetration and leverages the contact map constraint to optimize the partial poses only rather than the whole poses.

\section{Method}
Figure~\ref{fig.figure1} shows our method pipeline. It generates maps for contact areas by a network naming ContactCVAE, maps the contact maps to grasping poses by the other network naming GraspNet, and refines the generated grasp by a penetration-aware optimization module. In the work, we adopt MANO~\cite{romero2017embodied} to represent grasps. The MANO model $\mathcal{M}$ parameterizes the hand mesh $M=(V, F)$ ($V\in R^{778\times3}, F\in R^{1538\time 3}$ denotes the mesh vertices and faces) by the shape parameters $\beta\in R^{10}$ and pose parameters $\theta\in R^{51}$, \ie $M=\mathcal{M}(\theta,\beta)$. In the work, we use the mean shape and use $M=\mathcal{M}(\theta)$ for brevity.

In the first stage, ContactCVAE aims to learn a contact map distribution represented by a latent vector $z$ given an input object using a conditional variational autoencoder. The network takes an object point cloud $O\in R^{N\times3}$ and the contact map $C\in R^{N\times1}$ as the input and learns to make the output contact map $C^{'}\in R^{N\times1}$ as close to the input contact map as possible. $N$ is the number of points in $O$. Each point in the point cloud is represented by its normalized 3D positions. Each point in the contact map takes a value in $[0, 1]$ representing the contact score. During inference, given an object, a contact map $C^{'}$ can be generated by sampling from $z$. In the second stage, GraspNet learns a mapping from the contact map $C^{'}$ to the hand mesh $M$ constrained by the map. The pose $\theta'$ of the predicted mesh $M'$ from GraspNet is refined with PAP optimization in the third stage.

\subsection{ContactCVAE}
 Figure~\ref{fig.ContactCVAE} demonstrates the architecture of the ContactCVAE network, which is a generative model based on CVAE~\cite{sohn2015learning}. It consists of two blocks: a Condition-Encoder and a Generator. 
 
\noindent \textbf{Condition-Encoder} The Condition-Encoder \cgene is built on PointNet~\cite{qi2017pointnet}. It takes a point cloud as input to extract local features $f_{l}\in R^{N\times64}$ and global features$f_g\in R^{1\times1024}$. $f_g$ are then duplicated $N$ times to make a feature map $f_g\in R^{N\times 1024}$ for matching the shape of $f_{l}$. These two features are then concatenated as $f_{lg}$ for the conditional inputs for the generator below.
 
\noindent \textbf{Generator} The generator \gen follows an encoder-decoder architecture. As shown on the top of Figure~\ref{fig.ContactCVAE}, the encoder, \gene : $(C,O)\rightarrowtail z$, is based on PointNet~\cite{qi2017pointnet} architecture which takes both an object point cloud $O$ and a contact map $C$ as inputs and outputs the latent code $z\in R^{64}$. The encoder is only employed in training and is discarded in inference. The latent code $z$ represents a sample of the learned distribution $Q(z|\mu,\sigma^{2})$ and is used to generate the contact map, where $\mu,\sigma^{2}$ denotes the mean and variance of the distribution.  We then duplicate the latent code $z$ $N$ times to make the latent feature $f_z$ for all the points. The decoder \gend:$(f_z^i, f_{lg}^i)\rightarrowtail C'^i$ is a classifier for a point $i$ which merges three different features (global $f_g^i$, local $f_l^i$ and latent $f_z^i$) to classify whether the point belongs to a contact map or not. The decoder \gend uses the MLP architecture and the weights are shared for all points. 
\begin{figure*}[!htbp]
  \centering
  \vspace{-0.1cm}
\includegraphics[width=0.8\linewidth]{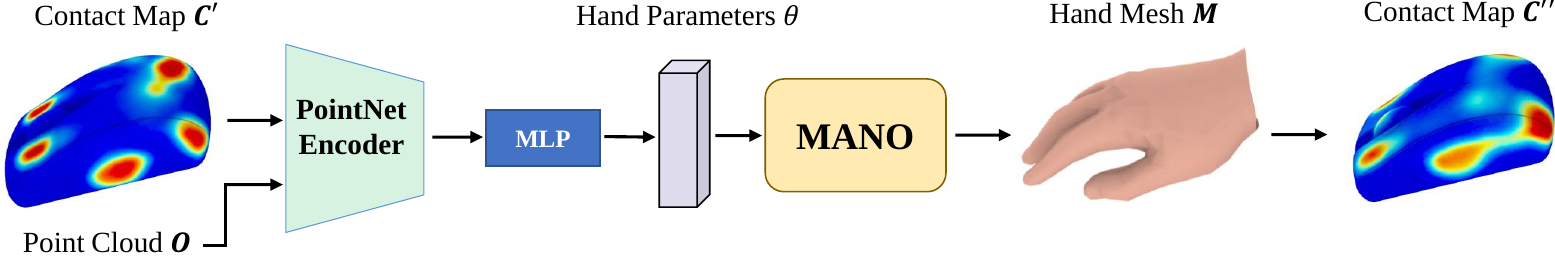}
  \caption{The architecture of GraspNet. It takes the concatenation of the generated (reconstructed) object contact $C'$ and the point cloud  $O$ as input to predict the grasp mesh parameterized by MANO.}
  \label{fig.graspnet}
    \vspace{-0.1cm}
\end{figure*}
\textbf{Testing Stage} During inference, as shown in the bottom of Figure~\ref{fig.ContactCVAE}, we only employ the Conditional-Encoder and decoder \gend. A latent code $z$ is randomly sampled from a Gaussian distribution and forms the latent feature $f_z$. At the same time, the Condition-Encoder takes an object point cloud to output the global and local features. With these features $(f_z,f_{g},f_{l})$, \gend outputs the grasp contact $C'$ for the object.

\noindent\textbf{Contact Loss} The goal of training the model is to optimize $\theta_e, \theta_d$ in order to reconstruct the contact map well. We simplify the goal as a binary classification task. Thus, we adopt the binary cross-entropy loss for the model over all the points, named as $L_{c1}$. However, some samples have small contact regions and it is hard for the model to learn those samples well by simply adopting the BCE loss. To address this problem, we additionally introduce the dice loss~\cite{milletari2016v} to train the model. It can assist the model in paying attention to small target regions. In our work, we adopt the dice loss for the 
same purpose and name as $L_{c2}$. the formulation of the two loss is defined as:
\begin{equation}
L_{c1} =\begin{matrix} -\sum_{i=0}^N[{y_i}log(\hat{y_i})+(1-y_i)log(1-\hat{y_i})], \end{matrix}
\end{equation}
\begin{equation}
L_{c2} =\begin{matrix} 1-\frac{2\sum_{i=0}^N y_i\hat{y_i}}{\sum_{i=0}^N y_i + \sum_{i=0}^N \hat{y_i}}, \end{matrix}
\end{equation}
where $\hat{y_i}$ and ${y_i}$ represent the predicted contact and ground truth of a point $i$, respectively. 

Following the training of CVAE~\cite{sohn2015learning}, we use the KL-Divergence loss regularizing the latent  distribution to be close to a standard Gaussian distribution. The loss term is named as $L_{kl}$.
The overall loss function of the ContactCVAE network, $L_{contact}$, is represented as:
\begin{equation}
    L_{contact} = \gamma_0L_{c1} +\gamma_1L_{c2} + \gamma_2L_{kl},
\end{equation}
where the $\gamma_0=0.5$, $\gamma_1=0 .5$ and $\gamma_2=1e-3$ are constants for balancing the loss terms.

\subsection{GraspNet}
With the assumption of hands full constrained by a contact map, we adopt a mapping function to get the grasping pose from the generated contact from the first stage.
As shown in Figure \ref{fig.graspnet}, the model takes an object point cloud $O$ and its generated (or reconstructed) contact $C'$ as the input to predict the hand mesh for the grasping pose, which is represented by the MANO model~\cite{romero2017embodied}. 
Specifically, we employ a PointNet~\cite{qi2017pointnet} to extract the feature and then use an MLP with four hidden layers to regress the MANO parameters. Given the parameters, the MANO model forms a differentiable layer that outputs the hand mesh $M$.  

During the training period, we use both ground truth and reconstructed contact maps to train the GraspNet. During inference, we only use the generated contact map to predict the grasp mesh. Both reconstructed and generated contact maps are from the ContactCVAE model in the first stage. 

\noindent\textbf{Reconstruction Loss} We simply adopt the reconstruction loss ($L_2$ distance) for the predicted vertices, named as $L_{v}$. The loss on MANO parameters is divided into two parts. We use the L1 loss for the translation parameter and the geodesic loss~\cite{mahendran20173d} for the pose parameter, named as $L_{t}$ and $L_{p}$ respectively. The final reconstruction error can be represented as $L_R = \lambda_vL_{v} + \lambda_tL_{t}  + \lambda_pL_{p}$, where $\lambda_v$=35,  $\lambda_t$=0.1 and $\lambda_p$=0.1 are constants balancing the losses. We also use the penetration loss $L_{ptr} = \frac{1}{\lvert O^{h}_{in}\rvert} \sum_{o\in O^{h}_{in}}{\min_{i}{\Vert o-V_i \Vert_2}}$ which punishes penetrations between the hand and object. $O^{h}_{in}$ denotes the object point subset that is inside the hand.

\noindent\textbf{Consistency Loss} Similar to the previous work~\cite{jiang2021hand}, we introduce the contact consistency loss $L_{cst} = \Vert C'-C''\Vert^2$. Based on the distance between the object and the grasp mesh $M$, the contact map $C''$ can be inferred by normalizing the distance between O and their nearest hand prior vertice. If the grasp mesh $M$ is predicted correctly from the GraspNet, the input contact map $C'$ should be consistent with the contact map $C''$.


The overall loss of GraspNet, $L_{grasp}$, is the weighted sum of all the above loss terms:
\begin{equation}
    L_{grasp} = L_R + \lambda_{ptr} L_{ptr} + \lambda_{cst}L_{cst},
\end{equation}
where $\lambda_{ptr}$=5 and $\lambda_{cst}$=0.05 denote the corresponding loss weights.

\begin{figure}[htbp]
  \centering
\includegraphics[width=0.8\linewidth]{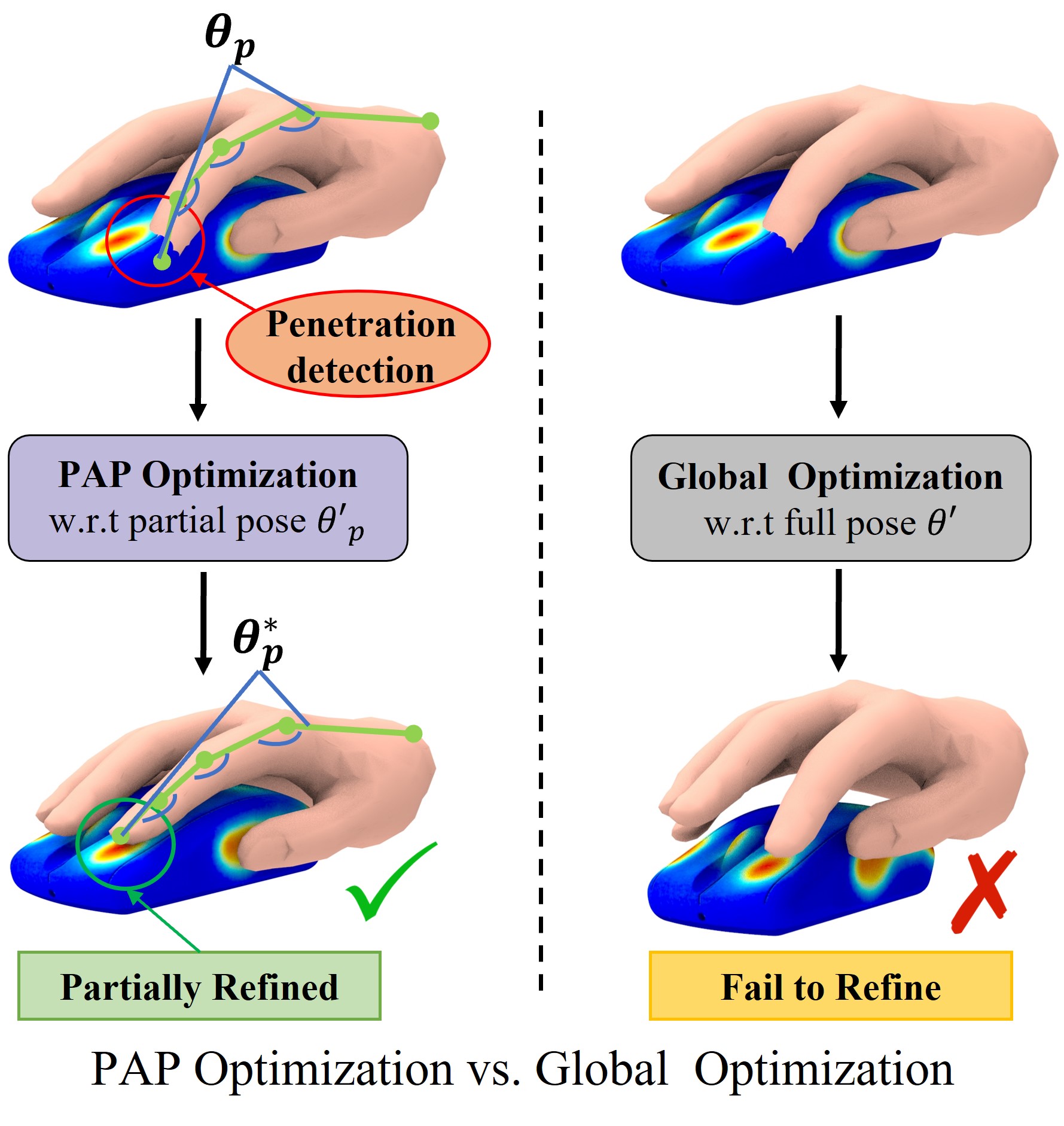}
  \caption{Left: illustration of our penetration-aware partial (PAP) optimization and the refined pose $\theta^*_p$. Penetration is detected in the index finger and therefore partial poses $\theta'_p$ on the index finger are to be optimized. Right: the refined pose of global optimization.}
  \label{fig.localrefine}
\vspace{-0.3cm}
\end{figure}

\begin{table*}[htbp]
    \vspace{-0.4cm}
    \centering
    \footnotesize
    \begin{tabular}{cclccccccc}
        \toprule
        \multirow{2}*{\textbf{Dataset}} &\multirow{2}*{\textbf{Methods}} & \multicolumn{2}{c}{\textbf{Ptr ($\downarrow$)}}  &\multicolumn{2}{c}{\textbf{SD ($\downarrow$})}
        &\multirow{2}*{\textbf{CR ($\uparrow$)}}
        &\multirow{2}*{\textbf{Div ($\uparrow$)}}
        &\multirow{2}*{\textbf{GSR ($\uparrow$)}}\\
        \cmidrule(r){3-4} \cmidrule(r){5-6} 
        &  &\textbf{Dep} &\textbf{Vol} &\textbf{Mean} &\textbf{Var}\\
        \midrule
          &GrabNet~\cite{taheri2020grab} &0.61 &8.31 &1.78 &$\pm$2.71  &98.25 &7.93 &27.60 \\
          &GraspField~\cite{karunratanakul2020grasping}  &0.56 &6.05 &2.07 &$\pm$2.81  &89.40 &- &-\\
        Obman &GraspTTA~\cite{jiang2021hand} &0.45 &5.14 &1.62 &\textbf{$\pm$2.18} &99.05 &8.07 &47.88\\

          &\textbf{Ours} &\textbf{0.44} &\textbf{3.94} &\textbf{1.60} &$\pm$2.28 &\textbf{100.00} &\textbf{10.14} &\textbf{61.37} \\
        \cmidrule(r){2-9}
          &GT  &0.01 &1.70 &1.66 &$\pm$2.44 &100.00  &7.86  &87.12 \\
          \midrule
           &GrabNet~\cite{taheri2020grab} &0.92 &16.74  &\textbf{1.04} &$\pm$1.60 &97.42 &5.92 &16.89 \\
          ContactPose &GraspTTA~\cite{jiang2021hand} &0.79 &6.01 &1.52 &$\pm$1.41 &97.67 &7.32 &42.31\\
            &\textbf{Ours} &\textbf{0.36} &\textbf{4.15} &1.40 &$\pm$1.98  &\textbf{98.85} &\textbf{7.91} &\textbf{58.97}\\
            \cmidrule(r){2-9}
            &GT &0.51 &5.91  &1.06 &$\pm$1.13 &99.65 &7.03 &41.78\\
        \bottomrule
    \end{tabular}
    \caption{Quantitative comparison with state-of-arts on Obman and ContactPose test set.}
    \tablelabel{table:ob_cp}
        \vspace{-0.2cm}
\end{table*}

\subsection{Penetration-aware Partial Optimization}
Though GraspNet gives plausible grasps for most cases, physically feasible grasps are sensitive to small errors in poses. For example, small penetration of a fingertip to the surface of an object can make the object drop to the floor. Hence, we propose the penetration-aware partial (PAP) optimization with generated contact maps to provide further constraints for small-scale partial pose refinement. PAP aims to detect the penetration and refine the partial poses causing it while keeping other partial poses of good quality unchanged. To this end, the full hand mesh is divided into six parts: five fingers and the palm. If penetration is detected in the palm area, all the poses are adjusted. If penetration is detected in a finger part and no penetration happens in the palm area, only the partial poses of the finger are adjusted. The loss for the PAP optimization is formulated as:

\begin{align}
    L_{opt}(\theta_p') =   & \omega_0L_{cst}(C'', C')  + \omega_1L_{ptr}(\mathcal{M}(\theta_p'), O ) + \nonumber \\
        +       & \omega_2L_{h}(\theta_p, \theta_p').
    \label{optim1}
\end{align}
$L_{cst}$, similar to the contact consistency loss defined above, is the difference between the generated contact map $C'$ and the contact map $C''$. $C''$ is obtained by normalizing the distance between the $O$ and their nearest point in $\mathcal{M}(\theta_p')$. $L_{ptr}$ penalizes the penetration between the hand $\mathcal{M}(\theta_p')$ and object $O$ as similar in \cite{jiang2021hand} and \cite{karunratanakul2020grasping}, defined above. $L_{h} =\Vert \theta_p - \theta_p'\Vert $  regularizes the pose hypothesis $\theta_p'$ to stay close to the generated pose $\theta_p$. We set $\omega_0$=0.1, $\omega_1$=2 and $\omega_2$=0.2.

Figure~\ref{fig.localrefine} (Left) shows an example of our partial optimization for the poses $\theta_p$ of the finger. During the optimization, as the loss mainly results from local wrong partial poses, the global optimization shown on the right side of Figure~\ref{fig.localrefine} has two issues 1) the gradient from local wrong partial poses affects other good poses, 2) the gradient cannot take full effect on the refinement for the wrong partial poses, which together results in many failures of small scale refinement. In contrast, our PAP only optimizes the poses causing the errors to get rid of these issues.

\section{Experiment}
\subsection{Implementation Details}
We sample $N=2048$ points on an object mesh as the input object point cloud. Our method is trained using a batch size of 32 examples, and an Adam  optimizer with a constant learning rate of 1e-4. The training dataset is randomly augmented with $[-1,1] cm$ translation and rotation at three (XYZ) dimensions. All the experiments were implemented in PyTorch, in which our models ran 130 epochs in a single RTX 3090 GPU with 24GB memory. In the Obman dataset~\cite{hasson2019learning}, all the ground truth contact map is derived by normalizing the distance between the ground truth of the hand and the object. For the inference refinement (both PAP and global optimization), the Adam optimizer with a learning rate of $2.0\times 10^{-4}$ is used. In the refinement process, each input is optimized for 200 steps.

\subsection{Datasets}
\textbf{Obman} We first validate our framework on the Obman dataset~\cite{hasson2019learning}, which is a large-scale synthetic dataset, including 3D hand interacting with objects. The hands are generated by a physical-based optimization engine GraspIt!~\cite{miller2004graspit}, and are parameterized by the MANO model~\cite{romero2017embodied}. The dataset contains 8 categories of everyday objects selected from ShapeNet~\cite{chang2015shapenet} with a total of 2772 meshes. The model trained on this dataset will benefit from the diversified object models. The object contact map is derived as \cite{taheri2020grab} by thresholding the normalized distance between the object points and their nearest hand vertices. Points with the distance smaller than a threshold are marked as contact points. 

\noindent\textbf{ContactPose} The ContactPose dataset~\cite{brahmbhatt2020contactpose} is a real dataset for studying hand-object interaction, which captures both ground-truth thermal contact maps and hand-object poses. Though the dataset contains only 25 household objects and 2306 grasp contacts, it captures more real interactions. For example, the contact in ContactPose spreads across large sections of the hand, as opposed to that at the fingertips for most cases in Obman. We manually split the dataset into a training and test group according to the object type. Specifically, we use 4 objects (cup, toothpaste, stapler, and flashlight) with 336 grasp contacts as a test set, and the rest for training the model. ContactPose uses the thermal camera-based method to capture the contact region.

\subsection{Evaluation Metrics}
A good generated pose should be physically stable and should be in contact with the object without penetration. In this work, we adopt three metrics to evaluate the quality of generated grasp poses: (1) \textbf{Penetration} (Ptr)  The penetration is measured by the depth (Dep, $cm$) and the volume (Vol, $cm^3$) between the objects and generated hand meshes. The depth is the maximum or mean of the distances from the hand mesh vertices to the surface of the object if penetration occurs. Following ~\cite{jiang2021hand,karunratanakul2020grasping}, the volume is measured by voxelizing the hand-object mesh with voxel size $0.5 cm$. (2) \textbf{Simulation Displacement} (SD) The simulation displacement is adopted to measure the stability of the generated grasp. We report the average (Mean, $cm$) and variance (Var, $cm$) of the simulation displacement as measured by a physics-based simulator following the same settings as ~\cite{jiang2021hand,karunratanakul2020grasping}. The displacement is the Euclidean distance between the object centers before and after applying a grasp on the object. Though used in the existing work, the results of previous work~\cite{karunratanakul2020grasping} indicate that a high penetration might correspond to a low simulation value and therefore we suggest readers use it for a rough reference only. (3) \textbf{Contact Rate} (CR, $\%$) A physically plausible hand-object interaction requires contact between the hand and the object. We define a sample as positive if the hand-object contact exists, which means that there exists at least a point on the hand surface is on or inside the surface of the object. The contact rate is the percentage of those positive samples over all the test samples. 

In addition to the metrics used in the hand generation work, we introduce two more metrics to evaluate the quality of the grasping pose distributions. (4) \textbf{Grasp Success Rate} (GSR, $\%$) The grasp success rate aims to evaluate the rate of grasp success. Specifically, we define the positive sample as the one with Ptr-Vol$<5 cm^3$ and SD-Mean $<2 cm$. The success rate is the percentage of those positive samples over all the test samples.
    (5) \textbf{Diversity} (Div, $cm$) It is also significant to evaluate the diversity for the generation task. In this work, we use MAE to measure the diversity of generated results. Specifically, we measure the divergence between each generated sample and all other samples and then average them. The formulation of the metrics mentioned above can be found in the supplementary material (Section B).

\begin{figure}[t]
  \centering
    \includegraphics[width=\linewidth]{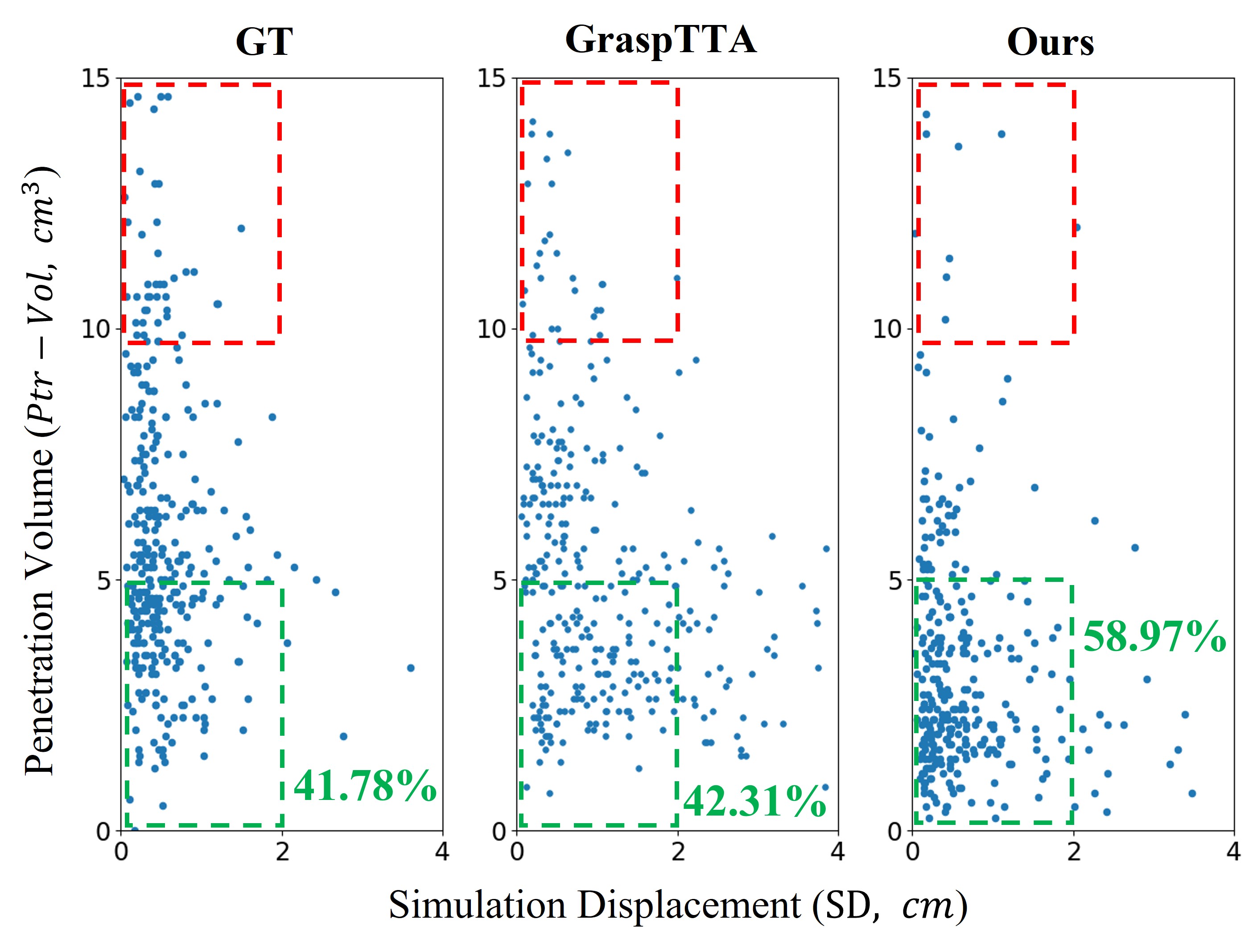}
  \caption{Scatter plots about metrics of Ptr-Vol vs. SD for the GT and sampled grasps of the ContactPose test set. Dots in the green box denote the positive samples grasping objects successfully during the simulation.}
  \label{fig.scatter_plot}
  \vspace{-0.4cm}
\end{figure}
\subsection{Comparison with state-of-arts}
To illustrate the advantages of the proposed method, we compared our method with three state-of-art methods: GrabNet~\cite{taheri2020grab}, GraspField~\cite{karunratanakul2020grasping} and GraspTTA~\cite{jiang2021hand}. We train the state-of-art methods on ContactPose using their public code, as they do not provide the results or metrics on this dataset. As for Obman, only the results of GraspField are quoted from~\cite{jiang2021hand}.

The results on Obman and ContactPose dataset in \tableref{table:ob_cp} show that our proposed method achieves the best performance on all the metrics.
Specifically, our method yields the best penetration depth (0.44$cm$ and 0.36$cm$) and volume (3.94$cm^3$ and 4.15$cm^3$). More importantly, our method achieves the best performance on diversity (10.14$cm$ and 7.91$cm$)  and GSR (61.37$\%$ and 58.97$\%$), indicating the robustness and variety of the results generated from our method. 

Note that due to the limitation of simulation, grasps with large penetration in many cases can still hold objects while a reasonable grasp pose should embody both low penetration and simulation displacement. Thus the GSR is a more comprehensive metric considering both of them. To verify this, Figure~\ref{fig.scatter_plot} plots penetration volume and simulation displacement of the GT grasps, sampled grasps from GraspTTA~\cite{jiang2021hand} and our method for the objects in the testing set of ContactPose. We can see that grasps in the red box of Figure~\ref{fig.scatter_plot} exhibiting larger penetration volume still demonstrate good grasp stability (small simulation displacement). Considering both metrics, the generated grasps from our method are closer to the origin, indicating the results have better stability with smaller penetration (see the comparisons of the green box).

 \begin{table}[!htbp]
    \centering
    \footnotesize
    \tabcolsep=0.8pt
    \begin{tabular}{clccccccc}
        \toprule
        \multirow{2}*{\textbf{Methods}} &\multicolumn{2}{c}{\textbf{Ptr}($\downarrow$)}  &\multicolumn{2}{c}{\textbf{SD ($\downarrow$)}}
        &\multirow{2}*{\textbf{CR ($\uparrow$)}}
        &\multirow{2}*{\textbf{Div ($\uparrow$)}}
        &\multirow{2}*{\textbf{GSR ($\uparrow$)}}\\
         \cmidrule(r){2-3} \cmidrule(r){4-5}
        &\textbf{Dep} &\textbf{Vol} &\textbf{Mean}  &\textbf{Var}\\
        \midrule
        Param2Mesh &1.02 &17.18 &1.14 &$\pm$1.80 &88.81 &6.31 &21.19\\
        Ours w/o PAP &0.60 &7.31 &\textbf{1.08} &\textbf{$\pm$1.18} &\textbf{98.85} &7.40 &35.71 \\
        Ours global opt &0.58 &6.67 &1.58 &$\pm$1.79 &96.23 &7.56 &39.27 \\
        Ours &\textbf{0.36} &\textbf{4.15} &1.40 &$\pm$1.98  &\textbf{98.85} &\textbf{7.91} &\textbf{58.97} \\
        \midrule
        Ours w/o PAP (GT) &0.56 &6.73  &0.98 &$\pm$0.97  &100.00 &6.79 &39.94 \\    
        Ours (GT) &0.40 &3.98  &1.06 &$\pm$1.13 &99.65 &7.03 &65.37 \\      

        \bottomrule
    \end{tabular}
    \caption{Self comparison on ContactPose test set.}
    \tablelabel{ablation}
    \vspace{-0.2cm}
\end{table}

\subsection{Ablation Study}

To verify the effectiveness of our proposed factorization and PAP optimization, we construct three variants of our method and a baseline, comparing their performances on ContactPose test set. The results are shown in \tableref{ablation}. \textbf{Param2Mesh}: A baseline for grasp generations learning an end-to-end grasp generation model. It learns the latent distribution of MANO parameter directly. 
Specifically, we use the same architecture of our ContactCVAE to make a fair comparison but replace the encoder \gene with fully connected layers to take the MANO parameters as inputs. Given a 3D object point cloud and a random sample for the distribution, the decoder \gend generates MANO parameters directly,  which is similar to the previous work~\cite{taheri2020grab}. \textbf{Ours w/o PAP}: A variant of our method removing the PAP optimization at the third stage. \textbf{Ours global opt}: A variant of our method simply adopting global refinement at the third stage. \textbf{Ours (GT)} skips ContactCVAE stage, training and testing GraspNet with GT contact maps directly, and \textbf{Ours w/o PAP (GT)} without the PAP optimization.

\begin{figure}[t]
  \centering
    \includegraphics[width=0.9\linewidth]{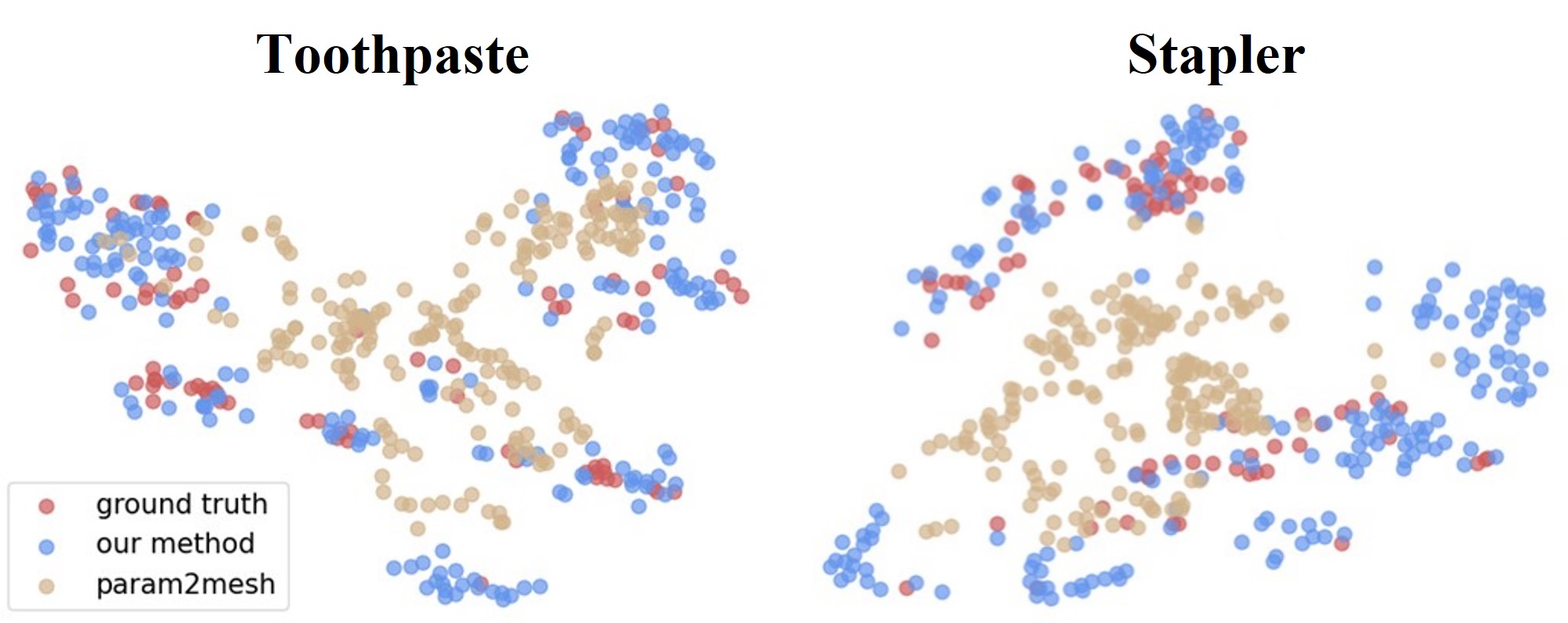}
  \caption{Visualization of the distribution of the generated grasp poses in the test objects (toothpaste and stapler) using t-SNE.}
  \label{fig.t-SNE}
  \vspace{-0.3cm}
\end{figure}

\noindent\textbf{Effectiveness of Factorization}
By comparing between Param2Mesh and Ours w/o PAP, we can see that our method achieves significant improvement on all metrics, indicating the effectiveness of the two-stage factorization. 
Especially, the penetration volume, depth and GSR are improved by 57$\%$, 38$\%$ and 68$\%$ respectively. In addition, the improvement on diversity (Div) indicates that our method generates more diverse samples. To further demonstrate the diversity and  generalization performance of our method, we select two test objects (toothpaste and stapler) of ContactPose  and visualize the distribution of generated grasp poses for each of them. As shown in Figure~\ref{fig.t-SNE}, we generate 200 grasp poses with small Ptr-Vol $(<5cm^3)$ and SD $(<2cm)$ for each object, and adopt the t-SNE technique to visualize the distribution of the pose parameters $\theta$. The sample distribution of our method is closer to the distribution of ground truth grasps, indicating better generalization performance. More results are shown in the supplementary material.

\noindent\textbf{Effectiveness of PAP Optimization} 
By comparing between Ours and Ours (global opt), we can see that our PAP optimization strategy achieves better performance than global refinement over all the metrics, which presents the effectiveness of our PAP optimization. 

\noindent\textbf{Quality of Generated Contact Maps}  
When compared with Our (GT) and Our w/o PAP (GT), the metrics for penetration of our methods (Ours and Our w/o PAP) are worse but the margin of corresponding improvement is relatively small. For example, the penetration depth of our methods are 0.36$mm$ and 0.60$mm$ while those of Our (GT) and Our w/o PAP (GT) are 0.40$mm$ and 0.56$mm$. The comparison indicates the generated maps in the first stage are of high quality. 

\begin{figure}[htbp]
  \centering
    \includegraphics[width=0.9\linewidth]{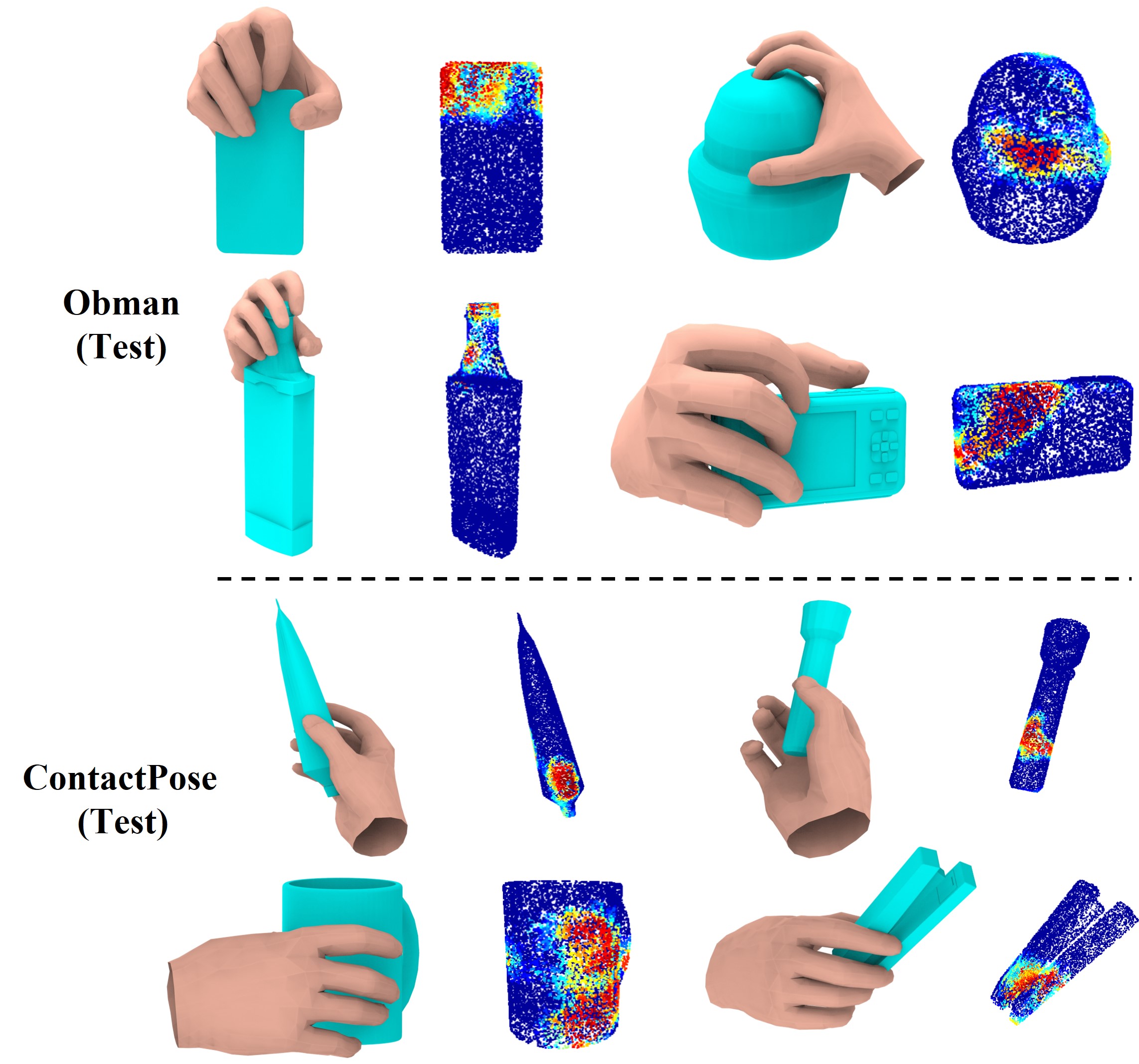}
  \caption{The visualization of generated contact maps and grasps for objects from the Obman test set and ContactPose test set. For each example, we present both the predicted grasp pose (left) and the corresponding contact map (right), which is presented in the form of the heat map.}
  \label{fig.qualitative result}
  \vspace{-0.4cm}
\end{figure}

The examples in Figure~\ref{fig.qualitative result} also show that the generated contacts convey meaningful information for grasping. We can observe that the generated contact map is reasonable, corresponding to the grasp pose. Although there are some failure examples (including unstable grasps and serious penetration), the hand pose is substantially natural as human behavior. 

\noindent\textbf{Semantic Analysis of Latent Contact Map Space} Using the generated object contacts to formulate the hand grasp is one of the contributions and here we show whether our ContactCVAE model can learn the latent distribution for contact well. In point generation work~\cite{achlioptas2018learning}, it demonstrates the quality of the generation model by showing that the learned representation is amenable to intuitive and semantically rich operations. Inspired by the work, we conduct the semantic analysis of the latent space learned from our ContactCVAE model. The detail of the procedure can be found in the supplementary material (Section D).

Figure \ref{fig.contact_analysis} exemplifies the contact maps and grasping poses generated by interpolating the latent $z$ of the contact maps for TypeA and TypeB. Notice that the grasps change gradually in alignment with the contact maps between Type A and Type B. 
For example, in Figure~\ref{fig.contact_analysis} (bottom), the yellow arrow and circle on the mouse, denote small differences between the contact maps and the grasp poses. As the contact region gradually appears, the middle finger moves to the corresponding position smoothly. Similar interesting observation can be found for manipulating the phone in Figure \ref{fig.contact_analysis} (top) where the hand poses change from holding to pressing gradually.

\section{Conclusion}
In this paper, we propose a novel framework for human grasps generation, which holds the potential for different deep architectures. The highlight of this work is exploiting the object affordance represented by the contact map, to formulate a functionality-oriented grasp pose and using penetration-aware partial optimization to refine partial-penetrated poses without hurting good-quality ones. The proposed method is extensively validated on two public datasets. In terms of diversity and stability, both quantitative and qualitative evaluations support that, our method has clear advantages over other strong competitors in generating high-quality human grasps. 

Although our method achieves great performance over all the metrics, limitations still exist. In some cases, the generated contact maps are ambiguous, resulting in more than one plausible grasping pose. Therefore, our assumption of the grasp is fully constrained by the contact map does not hold. Contact maps with more detailed hand-part segmentation can provide stronger constraints and help to reduce ambiguity.



\section*{Acknowledgments}
This work was supported in part by NSFC under Grants (No. 62103372, 62088101, and 62233013), PI funding of Zhejiang Lab (No. 121005-PI2101), Key Research Project of Zhejiang Lab (No. K2022PG1BB01), and OPPO Research Fund.
\bibliographystyle{named}
\bibliography{ijcai22}


\end{document}